\documentclass{article}

\usepackage[preprint]{neurips_2024}

\usepackage[utf8]{inputenc} 
\usepackage[T1]{fontenc}    
\usepackage{hyperref}       
\usepackage{url}            
\usepackage{booktabs}       
\usepackage{amsfonts}       
\usepackage{nicefrac}       
\usepackage{microtype}      
\usepackage{xcolor}         
\usepackage{graphicx}
\usepackage{amsmath}        

\title{DiscoverDCP: A Data-Driven Approach for Construction of Disciplined Convex Programs via Symbolic Regression}

\author{Sveinung Myhre \\
        Department of Electrical Engineering and Computer Sciences\\
        University of California, Berkeley \\
        Berkeley, CA 94720 \\
        \texttt{s.myhre@berkeley.edu} \\}
\begin{document}

\maketitle

\begin{abstract}
 We propose \textit{DiscoverDCP}, a data-driven framework that integrates symbolic regression with the rule sets of Disciplined Convex Programming (DCP) to perform system identification. By enforcing that all discovered candidate model expressions adhere to DCP composition rules, we ensure that the output expressions are globally convex by construction, circumventing the computationally intractable process of post-hoc convexity verification. This approach allows for the discovery of convex surrogates that exhibit more relaxed and accurate functional forms than traditional fixed-parameter convex expressions (e.g., quadratic functions). The proposed method produces interpretable, verifiable, and flexible convex models suitable for safety-critical control and optimization tasks.
\end{abstract}

\section{Introduction}

Convex optimization plays a central role in numerous applications including engineering, control theory, statistics, and machine learning, largely due to its strong theoretical guarantees for tractability and global optimality [1]. Ensuring that an optimization problem is convex relies on having a convex objective and convex constraints on a convex domain. Traditionally, practitioners rely on models known \textit{a priori} to be convex, such as linear or quadratic functions with positive definite Hessians, typically of the form:
\begin{equation}\label{eq:quadratic-expression}
   f(x) = x^\top A x + b^\top x + c.
\end{equation}
However, restricting models to such simple families can severely limit expressiveness and the ability to capture complex system dynamics.

The \textit{Disciplined Convex Programming} (DCP) framework, introduced by Grant et al. [2], provides a library of atoms and a set of compositional rules that guarantee convexity. If a function is constructed according to these rules (e.g., sums of convex functions, compositions with non-decreasing convex functions), the result is guaranteed to be convex. While the set of DCP-compliant functions is a subset of all convex functions, determining if a function is DCP is algorithmically straightforward [3], whereas verifying general convexity is NP-hard.

Despite the utility of DCP, discovering suitable convex expressions from raw data remains challenging. Existing practice typically involves linearization or fitting parametric convex families. In parallel, symbolic regression (SR) has emerged as a powerful technique for discovering interpretable analytic expressions directly from data [4]. While standard SR does not ensure convexity, modern tools allow for custom operator constraints.

This paper proposes \textit{DiscoverDCP}, a framework that leverages symbolic regression under the structural constraints imposed by DCP rules. By restricting the search space of symbolic regression to operations and compositions that preserve convexity, we algorithmically learn convex models from data. This approach yields interpretable mathematical expressions which are guaranteed to be convex by construction.

\paragraph{Our Contributions.}
To the best of our knowledge, this work represents the first direct integration of symbolic regression with Disciplined Convex Programming rulesets. Our specific contributions are:
\begin{itemize}
    \item We propose \textit{DiscoverDCP}, a method that restricts the symbolic regression search space to DCP-compliant operations, guaranteeing global convexity of learned models by construction.
    \item We demonstrate that this approach eliminates the need for intractable post-hoc convexity verification while offering greater interpretability than neural network-based approaches.
    \item We show via synthetic experiments that our method can better recover exact convex formulations where traditional quadratic baselines fail.
\end{itemize}

\section{Related Work}
\label{related_work}

Classic references for convex optimization include Boyd and Vandenberghe [1]. The notion of DCP [2] laid the groundwork for modern convex modeling toolboxes such as CVXPY [5], which automate the verification and solving of convex programs.

Symbolic regression has seen recent developments with high-performance tools like PySR [4]. While SR traditionally focuses on accuracy and parsimony, recent research has explored \textit{shape constraints}, such as ensuring positivity or monotonicity [6, 7]. 

In the realm of deep learning, Input Convex Neural Networks (ICNNs) [8] enforce convexity via constraints on weights and activation functions. However, ICNNs remain "black box" models that are difficult to interpret mathematically. In contrast, our proposed approach produces explicit, interpretable algebraic expressions that can be directly inspected and utilized in DCP-compatible solvers.

\section{Method}
\label{method}

The \textit{DiscoverDCP} framework integrates symbolic regression with the rules of Disciplined Convex Programming through two mechanisms:

\paragraph{1. Convexity-Preserving Operations.}  
DCP provides a set of atomic operations that preserve convexity. Our method restricts the symbolic search space to a subset of these atoms, specifically:
\begin{itemize}
    \item \textbf{Binary Operators:} Addition ($+$) (unrestricted), Multiplication ($*$) (restricted to constant scaling or specific convex-preserving forms), Maximum ($\max$).
    \item \textbf{Unary Operators:} Exponential ($\exp$), Square $(\cdot)^2$, and absolute value $|\cdot|$.
\end{itemize}
Crucially, the symbolic engine must enforce nesting rules. For example, a convex function can only be composed with a convex, non-decreasing function to remain convex [2].

\paragraph{2. Constrained Evolutionary Search.}
We employ PySR [4] as the search engine. The algorithm represents expressions as trees. We implement a constraint check at the mutation step of the genetic algorithm. If a mutation produces an expression tree that violates DCP rules (e.g., subtracting a convex function, or composing a convex function into a non-monotone function), the candidate is assigned an infinite loss or rejected. This ensures that the population of candidate models remains strictly within the set of DCP-compliant functions. A \textit{complexity measure} (number of nodes in the tree) penalizes overly convoluted expressions to maintain interpretability.

\section{Experiments}\label{experiments}

We evaluate the method on synthetic datasets where the ground-truth functions are known convex functions. This allows for controlled testing of whether \textit{DiscoverDCP} can recover the true structure or approximate it better than fixed parametric baselines.

Data points $x_i$ were sampled uniformly from a polytope domain defined by $Ax \leq b$, with Gaussian white noise added to the outputs. We compare our method against a standard quadratic baseline (Eq. \ref{eq:quadratic-expression}), where the matrix $A$ is constrained to be Positive Semidefinite (PSD) using CVXPY [5].

\begin{figure}[ht]
    \centering
    \includegraphics[width=0.9\linewidth]{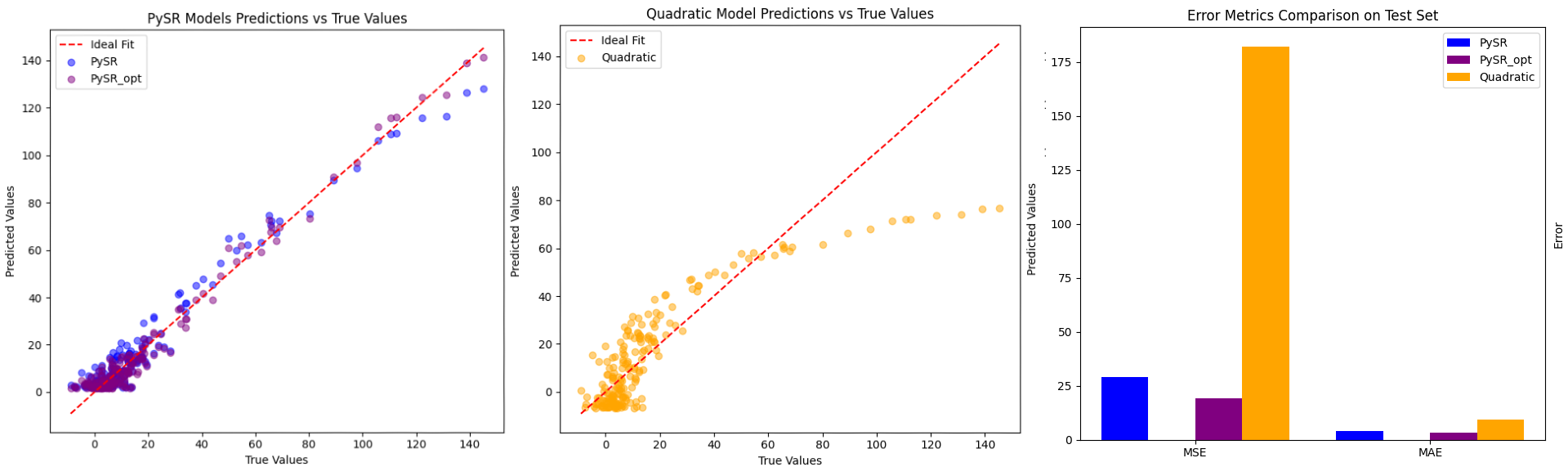}
    \caption{
    Comparison of model fits on 1D synthetic data. 
    \textbf{Red:} Ground truth ($y = \exp(x + \max(x, -5x) + x^{2}) + 4x$). 
    \textbf{Yellow:} PSD Quadratic fit ($y \approx 61.6 x^2 - 20.4x - 4.9$).
    \textbf{Blue:} \textit{DiscoverDCP} low-complexity approximation.
    \textbf{Purple:} \textit{DiscoverDCP} high-complexity approximation.
    }
    \label{fig:1d-pysr-example}
\end{figure}

\begin{figure}[ht]
    \centering
    \includegraphics[width=0.9\linewidth]{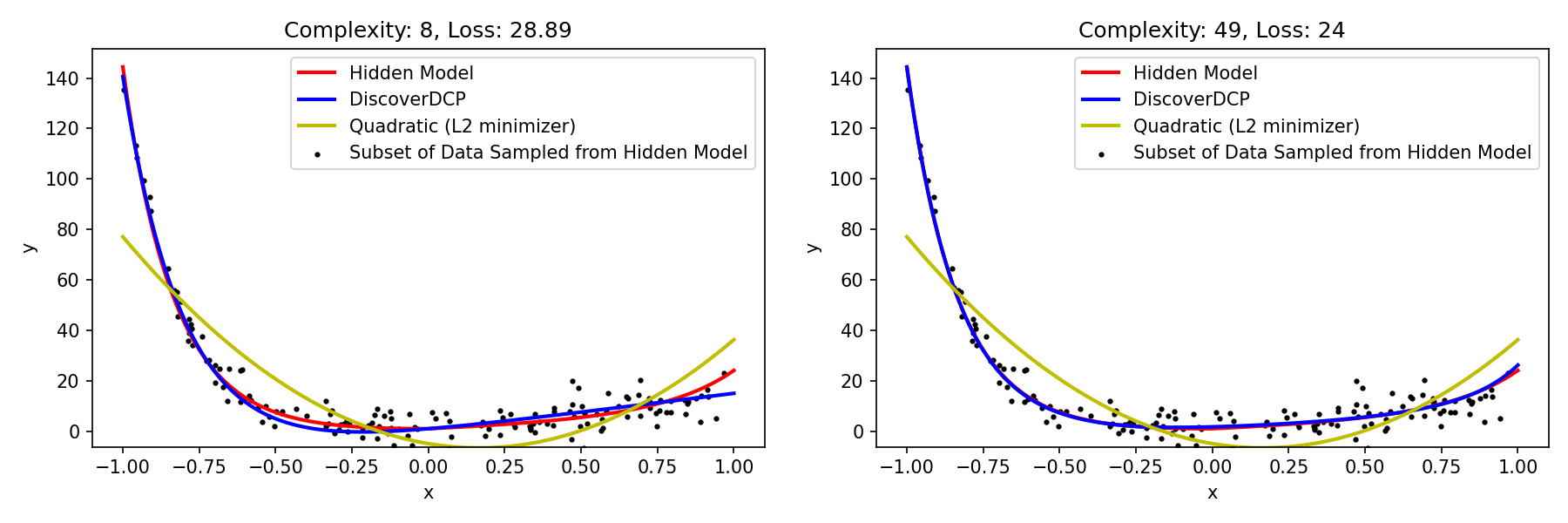}
    \caption{
    Functional forms of the discovered models. Left: A parsimonious approximation found by DiscoverDCP ($y=e^{3.2x} + e^{-4.9x}$). Right: A higher complexity model that accurately captures the nested non-linearities of the ground truth.
    }
    \label{fig:1d-pysr-example-objective-plots}
\end{figure}

Figures \ref{fig:1d-pysr-example} and \ref{fig:1d-pysr-example-objective-plots} illustrate the results on a hidden model with nested max and exponential terms. The quadratic baseline fails to capture the sharp non-linearities introduced by the `max` operator. In contrast, \textit{DiscoverDCP} finds a highly accurate model. 

While the quadratic model is convex, it lacks the flexibility to model the exponential growth and the switching behavior simultaneously. The high-complexity solution found by our method (Figure \ref{fig:1d-pysr-example-objective-plots}, Right) recovers a structure very similar to the ground truth, maintaining convexity guarantees throughout.

\section{Discussion}\label{discussion}

A key consideration in symbolic regression is the trade-off between accuracy and complexity. It is worth noting that a "simple" full quadratic expression (Eq. \ref{eq:quadratic-expression}) actually inhibits significant structural complexity as the dimension $n$ grows. The number of unique parameters in a PSD quadratic is:
\begin{equation}\label{eq:quadratic-complexity}
    P_{\text{quad}} = \frac{n (n+1)}{2} + n + 1.
\end{equation}
For high-dimensional systems, a quadratic model may have a higher parameter count than a sparse symbolic expression discovered by \textit{DiscoverDCP}. We propose that Eq. (\ref{eq:quadratic-complexity}) serves as a useful benchmark: if a symbolic model achieves lower loss than the quadratic baseline with a lower complexity score, it represents a superior representation of the convex system.

\section{Conclusion}\label{conclusion}

This paper introduced \textit{DiscoverDCP}, a framework that unifies symbolic regression with the structural rules of Disciplined Convex Programming (DCP). This combination allows the method to identify analytic expressions that are guaranteed to be convex by construction. The resulting models offer a verifiable and interpretable alternative to fixed-parameter approaches. Preliminary experiments indicate that \textit{DiscoverDCP} captures non-linear convex dynamics more effectively than traditional quadratic baselines.

\paragraph{Code Availability}
The code and experiments are available at: \url{https://github.com/svemyh/DiscoverDCP}.

\section*{References}

[1] Boyd, S., \& Vandenberghe, L. (2004). \textit{Convex Optimization}. Cambridge University Press.

[2] Grant, M., Boyd, S., \& Ye, Y. (2006). Disciplined Convex Programming. In \textit{Global Optimization}. Springer.

[3] Diamond, S., Chu, E., \& Boyd, S. (2013). DCP Analyzer. \url{https://dcp.stanford.edu/analyzer}

[4] Cranmer, M. (2023). Interpretable Machine Learning for Science with PySR and SymbolicRegression.jl. \textit{arXiv preprint arXiv:2305.01582}.

[5] Diamond, S., \& Boyd, S. (2016). CVXPY: A Python-embedded modeling language for convex optimization. \textit{Journal of Machine Learning Research}, 17(83), 1-5.

[6] Błądek, I., \& Krawiec, K. (2019). Solving Symbolic Regression Problems with Formal Constraints. In \textit{GECCO '19}.

[7] Aubin-Frankowski, P.-C., \& Szabo, Z. (2020). Hard Shape-Constrained Kernel Machines. \textit{NeurIPS 2020}.

[8] Amos, B., Xu, L., \& Kolter, J.Z. (2017). Input Convex Neural Networks. In \textit{ICML}.

\end{document}